# The Lovász-Bregman Divergence and connections to rank aggregation, clustering, and web ranking

**Rishabh Iyer**
Dept. of Electrical Engineering
University of Washington
Seattle, WA-98175, USA

**Jeff Bilmes**
Dept. of Electrical Engineering
University of Washington
Seattle, WA-98175, USA

## Abstract

We extend the recently introduced theory of Lovász Bregman (LB) divergences [19] in several ways. We show that they represent a distortion between a "score" and an "ordering", thus providing a new view of rank aggregation and order based clustering with interesting connections to web ranking. We show how the LB divergences have a number of properties akin to many permutation based metrics, and in fact have as special cases forms very similar to the Kendall-$\tau$ metric. We also show how the LB divergences subsume a number of commonly used ranking measures in information retrieval, like NDCG [22] and AUC [35]. Unlike the traditional permutation based metrics, however, the LB divergence naturally captures a notion of "confidence" in the orderings, thus providing a new representation to applications involving aggregating scores as opposed to just orderings. We show how a number of recently used web ranking models are forms of Lovász Bregman rank aggregation and also observe that a natural form of Mallow's model using the LB divergence has been used as conditional ranking models for the "Learning to Rank" problem.

## 1 Introduction

The Bregman divergence first appeared in the context of relaxation techniques in convex programming [5], and has found numerous applications as a general framework in clustering [3], proximal minimization ([7]), and others. Many of these applications are due to the nice properties of the Bregman divergence, and the fact that they are parameterized by a single convex function. They also generalize a large class of divergences between vectors.

In this paper, we investigate a specific class of Bregman divergences, parameterized via the Lovász extension of a submodular function. Submodular functions are a special class of discrete functions with interesting properties. Let $V$ refer to a finite ground set $\{1, 2, \ldots, |V|\}$. A set function $f : 2^V \to \mathbb{R}$ is submodular if $\forall S, T \subseteq V$, $f(S) + f(T) \geq f(S \cup T) + f(S \cap T)$. Submodular functions have attractive properties that make their exact or approximate optimization efficient and often practical [17, 21]. They also naturally arise in many problems in machine learning, computer vision, economics, operations research, etc. A link between convexity and submodularity is seen via the Lovász extension ([13, 29]) of the submodular function. While submodular functions are growing phenomenon in machine learning, recently there has been an increasing set of applications for the Lovász extension. In particular, recent work [1, 2] has shown nice connections between the Lovász extension and structured sparsity inducing norms.

This work is concerned with yet another application of the Lovász extension, in the form of the Lovász-Bregman divergence. This was first introduced in Iyer & Bilmes [19], in the context of clustering ranked vectors. We extend our work in several ways, mainly theoretically, by both showing a number of connections to the permutation based metrics, to rank aggregation, to rank based clustering and to the "Learning to Rank" problem in web ranking.

### 1.1 Motivation

The problems of rank aggregation and rank based clustering are ubiquitous in machine learning, information retrieval, and social choice theory. Below is a partial list of some of these applications.

**Meta Web Search:** We are given a collection of search engines, each providing a ranking or a score vector, and the task is to aggregate these to generate a combined result [27].

**Learning to Rank:** The "Learning to rank" problem, which is a fundamental problem in machine learning, involves constructing a ranking model from training data. This problem has gained significant interest in web ranking and information retrieval [28].

**Voter or Rank Clustering:** This is an important problem in social choice theory, where each voter provides a ranking or assigns a score to every item. A natural problem here is to meaningfully combine these rankings [26]. Sometimes however the population is heterogeneous and a mixture of distinct populations, each with its own aggregate representative, fits better.

**Combining Classifiers and Boosting:** There has been an increased interest in combining the output of different systems in an effort to improve performance of pattern classifiers, something often used in Machine Translation [34] and Speech Recognition[24]. One way of doing this [27] is to treat the output of every classifier as a ranking and combine the individual rankings of weak classifiers to obtain the overall classification. This is akin to standard boosting techniques [16], except that we consider rankings rather than just the valuations.

## 1.2 Permutation Based Distance Metrics

First a bit on notation – a permutation $\sigma$ is a bijection from $[n] = \{1, 2, \cdots, n\}$ to itself. Given a permutation $\sigma$, we denote $\sigma^{-1}$ as the inverse permutation such that $\sigma(i)$ is the item assigned rank $i$, while $\sigma^{-1}(j)$ is the rank[1] assigned to item $j$ and hence $\sigma(\sigma^{-1}(i)) = i$. We shall use $\sigma_x$ to denote a permutation induced through the ordering of a vector $x$ such that $x(\sigma_x(1)) \geq x(\sigma_x(2)) \cdots \geq x(\sigma_x(n))$. Without loss of generality, we assume that the permutation is defined via a decreasing order of elements. We shall use $v(i), v[i]$ and $v_i$ interchangeably to denote the $i$-th element in $v$. Given two permutations $\sigma, \pi$ we can define $\sigma\pi$ as the combined permutation, such that $\sigma\pi(i) = \sigma(\pi(i))$. Also given a vector $x$ and a permutation $\sigma$, define $x\sigma$ such that $x\sigma(i) = x(\sigma(i))$. We also define $\sigma x$ as $\sigma x(i) = x(\sigma^{-1}(i))$.

Recently a number of papers [27, 26, 23, 31] have addressed the problem of combining rankings using permutation based distance metrics. Denote $\mathbf{\Sigma}$ as the set of permutations over $[n]$. Then $d : \mathbf{\Sigma} \times \mathbf{\Sigma} \to \mathbb{R}_+$ is a permutation based distance metric if it satisfies the usual notions of a metric, viz. , $\forall \sigma, \pi, \tau \in \mathbf{\Sigma}, d(\sigma, \pi) \geq 0$ and $d(\sigma, \pi) = 0$ iff $\sigma = \pi$, $d(\sigma, \pi) = d(\pi, \sigma)$ and $d(\sigma, \pi) \leq d(\sigma, \tau) + d(\tau, \pi)$. In addition, to represent a distance amongst permutations, another property which is usually required is that of left invariance to reorderings, i.e., $d(\sigma, \pi) = d(\tau\sigma, \tau\pi)$[2]. The most standard notion of a permutation based distance metric is the Kendall $\tau$ metric [23]:

$$d_T(\sigma, \pi) = \sum_{i,j,i<j} I(\sigma^{-1}\pi(i) > \sigma^{-1}\pi(j)) \quad (1)$$

Where $I(.)$ is the indicator function. This distance metric represents the number of swap operations required to convert a permutation $\sigma$ to $\pi$. It's not hard to see that it is a metric and it satisfies the ordering invariance property. Other often used metrics include the Spearman's footrule $d_S$ and rank correlation $d_R$ [10]:

$$d_S(\sigma, \pi) = \sum_{i=1}^n |\sigma^{-1}(i) - \pi^{-1}(i)| \quad (2)$$

$$d_R(\sigma, \pi) = \sum_{i=1}^n (\sigma^{-1}(i) - \pi^{-1}(i))^2 \quad (3)$$

A natural extension to a ranking model is the Mallows model [30], which is an exponential model defined based on these permutation based distance metrics. This is defined as:

$$p(\pi|\theta, \sigma) = \frac{1}{Z(\theta)} \exp(-\theta d(\pi, \sigma)), \text{ with } \theta \geq 0. \quad (4)$$

This model has been generalized by [14] and also extended to multistage ranking by [15]. Lebanon and Lafferty [27] were amongst the first to use these models in machine learning by proposing an extended mallows model [14] to combine rankings in a manner like adaboost [16]. Similarly Meila *et al* [31] use the generalized Mallows model to infer the optimal combined ranking. Another related though different problem is clustering ranked data, investigated by [32], where they provide a $k$-means style algorithm. This was also extended to a machine learning context by [6].

## 1.3 Score based Permutation divergences

In this paper, we motivate another class of divergences, which capture the notion of distance between permutations. Unlike the permutation based distance metrics, however, these are distortion functions between a "score" and a permutation. This, as we shall see, offers a new view of rank aggregation and order based clustering problems. We shall also see a number of interesting connections to web ranking.

Consider a scenario where we are given a collection of scores $x^1, x^2, \cdots, x^n$ as opposed to just a collection of orderings – i.e., each $x^i$ is an ordered vector and not just a permutation. This occurs in a number of real world applications. For example, in the application of combining classifiers [27], the classifiers often output scores (in the form of say normalized confidence or probability distributions). While the rankings themselves are informative, it is often more beneficial to use the additional information in the form of scores

---

[1]This is opposite from the convention used in [27, 26, 23, 31] but follows the convention of [17].

[2]While in the literature this is called right invariance, we have left invariance due to our notation

if available. This in some sense combines the approach of Adaboost [16] and Cranking [27], since the former is concerned only with the scores while the latter takes only the orderings. The case of voting is similar, where each voter might assign scores to every candidate (which can sometimes be easier than assigning an ordering). This also applies to web-search where often the individual search engines (or possibly features) provide a confidence score for each webpage. Since these applications provide both the valuations and the rankings, we call these *score based ranking applications*.

A *score based permutation divergence* is defined as follows. Given a convex set $\mathbb{S}$, denote $d : \mathbb{S} \times \mathbf{\Sigma} \to \mathbb{R}_+$ as a score based permutation divergence if $\forall x \in \mathbb{S}, \sigma \in \mathbf{\Sigma}, d(x||\sigma) \geq 0$ and $d(x||\sigma) = 0$ if and only if $\sigma_x = \sigma$. Another desirable property is that of left invariance, viz. $d(x||\sigma) = d(\tau x||\tau \sigma), \forall \tau, \sigma \in \mathbf{\Sigma}, x \in \mathbb{S}$.

It is then immediately clear how the score based permutation divergence naturally models the above scenario. The problem becomes one of finding a representative ordering, i.e., find a permutation $\sigma$ that minimizes the average distortion to the set of points $x^1, \cdots, x^n$. Similarly, in a clustering application, to cluster a set of ordered scores, a score based permutation divergence fits more naturally. The representatives for each cluster are permutations, while the objects themselves are ordered vectors. Notice that in both cases, a purely permutation based distance metric would completely ignore the values, and just consider the induced orderings or permutations. To our knowledge, this work is the first time that the notion of a score based permutation divergence has been introduced formally, thus providing a novel view to the rank aggregation and rank based clustering problems.

### 1.4 Our Contributions

In this paper, we investigate several theoretical properties of one such score based permutation divergence – the LB divergence. This work builds on our previous work [19], where we introduce the Lovász-Bregman divergence. Our focus therein is mainly on the connections between the Lovász Bregman and a discrete Bregman divergence connected with submodular functions and we also provide a k-means framework for clustering ordered vectors. In the present paper, however, we make the connection to rank aggregation and clustering more precise, by motivating the class of score based permutation divergences and showing relations to permutation based metrics and web ranking.

The following are some of our main results. We introduce a novel notion of the generalized Bregman divergence based on a "subgradient map". While this is of independent theoretical interest, it helps us characterize the Lovász-Bregman divergence. We then show that the LB divergence is indeed a score based permutation divergence with several similarities to permutation based metrics. In fact, we show that a form of weighted Kendall $\tau$, and a form related to the Spearman's Footrule, occurs as instances of the Lovász-Bregman divergences. We also show how a number of loss functions used in IR and web ranking like the Normalized Discounted Cumulative Gain (NDCG) [22] and the Area Under the Curve (AUC) [35] occur as instances of the LB. We then demonstrate some unique properties of the Lovász Bregman divergence not present in permutation-distance metrics. Notable amongst these are the properties that the Lovász-Bregman naturally captures a notion of "confidence" of an ordering, and exhibits a priority for higher rankings, both of which are desirable in score based ranking applications. We then define a Lovász-Mallows model as a conditional model over both the scores and the ranking. We finally connect the Lovász Bregman to rank aggregation and rank based clustering. We show in fact that a number of ranking models for web ranking used in the past are instances of Lovász Bregman rank aggregation. We moreover show that a number of conditional models used in the past for learning to Rank are closely related to the Lovász-Mallows model.

## 2 The Lovász Bregman divergences

In this section, we shall briefly review the Lovász extension and define forms of the generalized Bregman and the LB divergence. We only state the main results here and for a more extensive discussion, refer to [20].

### 2.1 The Generalized Bregman divergences

The notion used in this section follows from [33, 37]. We denote $\phi$ as a proper convex function (i.e., it's domain is non-empty and it does not take the value $-\infty$), reint(.) and dom(.) as the relative interior and domain respectively. A subgradient $g$ at $y \in \text{dom}(\phi)$ is such that for any $x, \phi(x) \geq \phi(y) + \langle g, x-y \rangle$ and the set of all subgradients at $y$ is the subdifferential and is denoted by $\partial_\phi(y)$.

The Taylor series approximation of a twice differentiable convex function provides a natural way of generating a Bregman divergence ([5]). Given a twice differentiable and strictly convex function $\phi$:

$$d_\phi(x, y) = \phi(x) - \phi(y) - \langle \nabla \phi(y), x - y \rangle. \quad (5)$$

In order to extend this notion to non-differentiable convex functions, generalized Bregman divergences have been proposed [37, 25]. While gradients no longer exist at points of non-differentiability, the directional derivatives exist in the relative interior of the domain of $\phi$, as long as the function is finite. Hence a natural formulation is to replace the gradient by the directional derivative, a notion which has been pursued in [37, 25].

In this paper, we view the generalized Bregman divergences slightly differently, in a way related to the approach in [18]. In order to ensure that the subgradients exist, we only consider the relative interior of the domain. Then define $\mathcal{H}_\phi(y)$ as a subgradient-map such that $\forall y \in \text{reint}(\text{dom}(\phi)), \mathcal{H}_\phi(y) \in \partial_\phi(y)$. Then given $x \in \text{dom}(\phi), y \in \text{reint}(\text{dom}(\phi))$ and a subgradient map $\mathcal{H}_\phi$, we define the generalized Bregman divergence as:

$$d_\phi^{\mathcal{H}_\phi}(x,y) = \phi(x) - \phi(y) - \langle \mathcal{H}_\phi(y), x - y \rangle \quad (6)$$

When $\phi$ is differentiable, notice that $\partial_\phi(y) = \{\nabla \phi(y)\}$ and hence $\mathcal{H}_\phi(y) = \nabla(y)$.

## 2.2 Properties of the Lovász Extension

We review some important theoretical properties of the Lovász extension. Given any vector $y \in [0,1]^n$ and it's associated permutation $\sigma_y$, define $S_j^{\sigma_y} = \{\sigma_y(1), \cdots, \sigma_y(j)\}$ for $j \in [n]$. Notice that in general $\sigma_y$ need not be unique (it will be unique only if $y$ is totally ordered), and hence let $\boldsymbol{\Sigma}_y$ represent the set of all possible permutations with this ordering. Then the Lovász extension of $f$ is defined as:

$$\hat{f}(y) = \sum_{j=1}^n y[\sigma_y(j)](f(S_j^{\sigma_y}) - f(S_{j-1}^{\sigma_y})) \quad (7)$$

This is also called the Choquet integral [9] of $f$. Though $\sigma_y$ might not be unique, the Lovász extension is actually unique. Furthermore, $\hat{f}$ is convex if and only if $f$ is submodular. In addition, the Lovász extension is also tight on the vertices of the hypercube, in that $f(X) = \hat{f}(1_X), \forall X \subseteq V$ (where $1_X$ is the characteristic vector of $X$, i.e., $1_X(j) = I(j \in X)$) and hence it is a valid continuous extension. The Lovász extension is in general a non-smooth convex function, and hence there does not exist a unique subgradient at every point. The following result due [17, 13] provides a characterization of the extreme points of the Lovász subdifferential polyhedron $\partial \hat{f}(y)$:

**Lemma 2.1.** *[17, 13] For a submodular function $f$, a vector $y$ and a permutation $\sigma_y \in \boldsymbol{\Sigma}_y$, a vector $h_{\sigma_y}^f$ defined as:*

$$h_{\sigma_y}^f(\sigma_y(j)) = f(S_j^{\sigma_y}) - f(S_{j-1}^{\sigma_y}), \forall j \in \{1, 2, \cdots, n\}$$

*forms an extreme point of $\partial \hat{f}(y)$. Also, the number of extreme points of $\partial \hat{f}(y)$ is $|\boldsymbol{\Sigma}_y|$.*

Notice that the extreme subgradients are parameterized by the permutation $\sigma_y$ and hence we refer to them as $h_{\sigma_y}^f$. Seen in this way, the Lovász extension then takes an extremely simple form: $\hat{f}(w) = \langle h_{\sigma_w}^f, w \rangle$.

We now point out an interesting property related to the extreme subgradients of $\hat{f}$. Define $\mathcal{P}(\sigma)$ as a $n$-simplex corresponding to a permutation $\sigma$ (or chain $\mathcal{C}^\sigma : \emptyset \subset S_1^\sigma \subset \cdots \subset S_n^\sigma = V$). In other words, $\mathcal{P}(\sigma) = \text{conv}(1_{S_i^\sigma}, i = 1, 2, \cdots, n)$. It's easy to see that $\mathcal{P}(\sigma) \subseteq [0,1]^n$.

**Lemma 2.2.** *(Lemma 6.19, [17]) Given a permutation $\sigma \in \boldsymbol{\Sigma}$, for every vector $y \in \mathcal{P}(\sigma)$ the vector $h_\sigma^f$ is an extreme subgradient of $\hat{f}$ at $y$. If $y$ belongs to the (strict) interior of $\mathcal{P}(\sigma)$, $h_\sigma^f$ is a unique subgradient corresponding to $\hat{f}$ at $y$.*

The above lemma points out a critical fact about the subgradients of the Lovász extension, in that they depend only on the total ordering of a vector and are independent of the vector itself. This also implies that if $y$ is totally ordered (it belongs to the interior of $\mathcal{P}(\sigma_y)$) then $\partial_{\hat{f}}(y)$ consists of a single (unique) subgradient. Hence, two entirely different but identically ordered vectors will have identical extreme subgradients. This fact is important when defining and understanding the properties of the LB divergence.

## 2.3 The Lovász Bregman divergences

We are now in position to define the Lovász-Bregman divergence. Throughout this paper, we restrict $\text{dom}(\hat{f})$ to be $[0,1]^n$. For the applications we consider in this paper, we lose no generality with this assumption, since the scores can easily be scaled to lie within this volume.

Consider the case when $y$ is totally ordered, and correspondingly $|\boldsymbol{\Sigma}_y| = 1$. It follows then from Lemma 2.2 that there exists a unique subgradient and $\mathcal{H}_{\hat{f}}(y) = h_{\sigma_y}^f$. Hence for any $x \in [0,1]^n$, we have from Eqn. (6) that [19]:

$$d_{\hat{f}}(x,y) = \hat{f}(x) - \langle x, h_{\sigma_y}^f \rangle = \langle x, h_{\sigma_x}^f - h_{\sigma_y}^f \rangle \quad (8)$$

Notice that this divergence depends only on $x$ and $\sigma_y$, and is independent of $y$ itself. In particular, the LB divergence between a vector $x$ and any vector $y \in \mathcal{P}(\sigma)$ is the same for all $y \in \mathcal{P}(\sigma)$ (Lemma 2.2). We also invoke the following lemma from [19]:

**Lemma 2.3.** *(Theorem 2.2, [19]) Given a submodular function whose polyhedron contains all possible extreme points and $x$ which is totally ordered, $d_{\hat{f}}(x,y) = 0$ if and only if $\sigma_x = \sigma_y$.*

At first sight it seems that the class of submodular functions satisfying Lemma 2.3 is very specific. We point out however that this class is quite general and many instances we consider in this paper belong to this class of functions. For example, it is easy to see that the class of submodular functions $f(X) = g(|X|)$ where $g$ is a concave function satisfying $g(i) - g(i-1) \neq g(j) - g(j-1)$ for $i \neq j$ belong to this class.

Hence the Lovász-Bregman divergence is score based permutation divergence, and we denote it as:

$$d_{\hat{f}}(x||\sigma) = \langle x, h_{\sigma_x}^f - h_\sigma^f \rangle \quad (9)$$

| | $f(X)$ | $\hat{f}(x)$ | $d_{\hat{f}}(x, y)$ |
|---|---|---|---|
| 1) | $\|X\|\|V \setminus X\|$ | $\sum_{i<j} \|x_i - x_j\|$ | $\sum_{i<j} \|x_{\sigma(i)} - x_{\sigma(j)}\| I(\sigma_x^{-1}\sigma(i) > \sigma_x^{-1}\sigma(j))$ |
| 2) | $g(\|X\|)$ | $\sum_{i=1}^k x(\sigma_x(i))\delta_g(i)$ | $\sum_{i=1}^n x(\sigma_x(i))\delta_g(i) - \sum_{i=1}^k x(\sigma_y(i))\delta_g(i)$ |
| 3) | $\min\{\|X\|, k\}$ | $\sum_{i=1}^k x(\sigma_x(i))$ | $\sum_{i=1}^k x(\sigma_x(i)) - \sum_{i=1}^k x(\sigma(i))$ |
| 4) | $\min\{\|X\|, 1\}$ | $\max_i x_i$ | $\max_i x_i - x(\sigma(1))$ |
| 5) | $\sum_{i=1}^n \|I(i \in X) - I(i+1 \in X)\|$ | $\sum_{i=1}^n \|x_i - x_{i+1}\|$ | $\sum_{i=1}^n \|x_i - x_{i+1}\| I(\sigma_x^{-1}\sigma(i) > \sigma_x^{-1}\sigma(i+1))$ |
| 6) | $I(1 \leq \|A\| \leq n-1)$ | $\max_i x(i) - \min_i x(i)$ | $\max_i x(i) - x(\sigma(1)) - \min_i x(i) + x(\sigma(n))$ |
| 7) | $I(A \neq \emptyset, A \neq V)$ | $\max_{i,j} \|x_i - x_j\|$ | $\max_{i,j} \|x_i - x_j\| - \|x(\sigma(1) - x(\sigma(n))\|$ |

Table 1: Examples of the LB divergences. $I(.)$ is the Indicator fn.

As we shall see in the next section, this divergence has a number of properties akin to the standard permutation based distance metrics. Since a large class of submodular functions satisfy the above property (of having all possible extreme points), the Lovász-Bregman divergence forms a large class of divergences.

The case when $y$ is not totally ordered can be handled similarly [20].

### 2.4 Lovász Bregman Divergence Examples

Below is a partial list of some instances of the Lovász-Bregman divergence. We shall see that a number of these are closely related to many standard permutation based metrics. Table 1 considers several other examples of LB divergences.

**Cut function and symmetric submodular functions:** A fundamental submodular function, which is also symmetric, is the graph cut function. This is $f(X) = \sum_{i \in X} \sum_{j \in V \setminus X} d_{ij}$. The Lovász extension of $f$ is $\hat{f}(x) = \sum_{i,j} d_{ij}(x_i - x_j)_+$ [2]. The LB divergence corresponding to $\hat{f}$ then has a nice form:

$$d_{\hat{f}}(x\|\sigma) = \sum_{i<j} d_{\sigma(i)\sigma(j)}|x_{\sigma(i)} - x_{\sigma(j)}|I(\sigma_x^{-1}\sigma(i) > \sigma_x^{-1}\sigma(j)) \quad (10)$$

We in addition assume that $d$ is symmetric (i.e., $d_{ij} = d_{ji}, \forall i, j \in V$) and hence $f$ is also symmetric. Indeed a weighted version of Kendall $\tau$ can be written as $d_T^w(\sigma, \pi) = \sum_{i,j:i<j} w_{ij}I(\sigma^{-1}\pi(i) > \sigma^{-1}\pi(j))$ and $d_{\hat{f}}(x\|\sigma)$ is exactly then a form of $d_T^w(\sigma_x, \sigma)$, with $w_{ij} = d_{\sigma(i)\sigma(j)}|x_{\sigma(i)} - x_{\sigma(j)}|$. Moreover, if $d_{ij} = \frac{1}{|x_i - x_j|}$, we have $d_{\hat{f}}(x\|\sigma) = d_T(\sigma_x, \sigma)$. Hence, we recover the Kendall $\tau$ for that particular $x$.

An interesting special case of this is when $f(X) = \|X\|\|V \setminus X\|$, in which case we get:

$$d_{\hat{f}}(x\|\sigma) = \sum_{i<j} |x_{\sigma(i)} - x_{\sigma(j)}| I(\sigma_x^{-1}\sigma(i) > \sigma_x^{-1}\sigma(j)).$$

**Cardinality based monotone submodular functions:** Another class of submodular functions is $f(X) = g(\|X\|)$ for some concave function $g$. This form induces an interesting class of Lovász Bregman divergences. In this case $h_{\sigma_x}^f(\sigma_x(i)) = g(i) - g(i-1)$. Define $\delta_g(i) = g(i) - g(i-1)$, then:

$$d_{\hat{f}}(x\|\sigma) = \sum_{i=1}^n x[\sigma_x(i)]\delta_g(i) - \sum_{i=1}^n x[\sigma(i)]\delta_g(i). \quad (11)$$

Notice that we can start with any $\delta_g$ such that $\delta_g(1) \geq \delta_g(2) \geq \cdots \geq \delta_g(n)$, and through this we can obtain the corresponding function $g$. Consider a specific example, with $\delta_g(i) = n - i$. Then, $d_{\hat{f}}(x\|\sigma) = \sum_{i=1}^n x[\sigma(i)]i - x[\sigma_x(i)]i = \langle x, \sigma^{-1} - \sigma_x^{-1}\rangle$. This expression looks similar to the Spearman's rule (Eqn. (2)), except for being additionally weighted by $x$.

We can also extend this in several ways. For example, consider a restriction to the top $m$ elements ($m < n$). Define $f(X) = \min\{g(\|X\|), g(m)\}$. Then it is not hard to verify that:

$$d_{\hat{f}}(x\|\sigma) = \sum_{i=1}^m x[\sigma_x(i)]\delta_g(i) - \sum_{i=1}^m x[\sigma(i)]\delta_g(i). \quad (12)$$

A specific example is $f(X) = \min\{\|X\|, m\}$, where

$$d_{\hat{f}}(x\|\sigma) = \sum_{i=1}^m x(\sigma_x(i)) - x(\sigma(i)). \quad (13)$$

In this case, the divergence between $x$ and $\sigma$ is the difference between the largest $m$ values of $x$ and the $m$ first values of $x$ under the ordering $\sigma$. Here the ordering is not really important, but it is just the sum of the top $m$ values and hence if $\sigma_x$ and $\sigma$, under $x$, have the same sum of first $m$ values, the divergence is zero (irrespective of their ordering or individual element valuations). We can also define $\delta_g$, such that $\delta_g(1) = 1$ and $\delta_g(i) = 0, \forall i \neq 1$. Then, $d_{\hat{f}}(x\|\sigma) = \max_j x(j) - x(\sigma(1))$ (this is equivalent to Eqn. (13) when $m = 1$). In this case, the divergence depends only on the top value, and if $\sigma_x$ and $\sigma$ have the same leading element, the divergence is zero.

### 2.5 Lovász Bregman as ranking measures

In this section, we show how the Lovász Bregman subsumes and is closely related to several commonly used loss functions in Information Retrieval connected to ranking.

**The Normalized Discounted Cumulative Gain (NDCG):** The NDCG metric [22] is one of the most widely used ranking measures in web search. Given a relevance vector $r$, where the entry $r_i$ typically provides the relevance of a document $i \in \{1, 2, \cdots, n\}$ to a query, and an ordering of documents $\sigma$, the NDCG loss function with respect to a discount function $D$ is defined as:

$$\mathcal{L}(\sigma) = \frac{\sum_{i=1}^{k} r(\sigma_r(i))D(i) - \sum_{i=1}^{k} r(\sigma(i))D(i)}{\sum_{i=1}^{k} r(\sigma_r(i))D(i)} \quad (14)$$

Here $k \leq n$ is often used as a cutoff. Intuitively the NDCG loss compares an ordering $\sigma$ to the best possible ordering $\sigma_r$. The typical choice of $D(i) = \frac{1}{\log(1+i)}$, though in general any decreasing function can be used. This function is closely related to a form of the LB divergence. In particular, notice that $\mathcal{L}(\sigma) \propto \sum_{i=1}^{k} r(\sigma_r(i))D(i) - \sum_{i=1}^{k} r(\sigma(i))D(i)$ (since the denominator of Eqn. (14) is a constant) which is form of Eqn. (12) with $m = k$ and choosing the function $g(i) = \sum_{j=1}^{i} D(i)$.

**Area Under the Curve:** Another commonly used ranking measure is the Area under the curve [35]. Unlike NDCG however, this just relies on a partial ordering of the documents and not a complete ordering. In particular denote $G$ as a set of "good" documents and $B$ as a set of "bad" documents. Then the loss function $\mathcal{L}(\sigma)$ corresponding to an ordering of documents $\sigma$ is

$$\mathcal{L}(\sigma) = \frac{1}{|G||B|} \sum_{g \in G, b \in B} I(\sigma(g) > \sigma(b)). \quad (15)$$

This can be seen as an instance of LB divergence corresponding to the cut function by choosing $d_{ij} = \frac{1}{|G||B|}, \forall i, j, x_g = 1, \forall g \in G$ and $x_b = 0, \forall b \in B$.

## 3 Lovász Bregman Properties

In this section, we shall analyze some interesting properties of the LB divergences. While many of these properties show strong similarities with permutation based metrics, the Lovász Bregman divergence enjoys some unique properties, thereby providing novel insight into the problem of combining and clustering ordered vectors.

**Non-negativity and convexity:** The LB divergence is a divergence, in that $\forall x, \sigma, d_{\hat{f}}(x||\sigma) \geq 0$. Additionally if the submodular polyhedron of $f$ has all possible extreme points, $d_{\hat{f}}(x||\sigma) = 0$ iff $\sigma_x = \sigma$. Also the Lovász-Bregman divergence $d_{\hat{f}}(x||\sigma)$ is convex in $x$ for a given $\sigma$.

**Equivalence Classes:** The LB divergence of submodular functions which differ only in a modular term are equal. Hence for a submodular function $f$ and a modular function $m$, $d_{\widehat{f+m}}(x||\sigma) = d_{\hat{f}}(x||\sigma)$. Since any submodular function can be expressed as a difference between a polymatroid and a modular function [11], it follows that it suffices to consider polymatroid functions while defining the LB divergences.

**Linearity and Linear Separation:** The LB divergence is a linear operator in the submodular function $f$. Hence for two submodular functions $f_1, f_2, d_{\widehat{f_1+f_2}}(x||\sigma) = d_{\hat{f}_1}(x||\sigma) + d_{\hat{f}_2}(x||\sigma)$. The LB divergence has the property of linear separation — the set of points $x$ equidistant to two permutations $\sigma_1$ and $\sigma_2$ (i.e., $\{x : d_{\hat{f}}(x||\sigma_1) = d_{\hat{f}}(x||\sigma_2)\}$) comprise a hyperplane. Similarly, for any $x$, the set of points $y$ such that $d_{\hat{f}}(x, y) = $ constant, is $\mathcal{P}(\sigma_y)$.

**Invariance over relabelings:** The permutation based distance metrics have the property of being left invariant with respect to reorderings, i.e., given permutations $\pi, \sigma, \tau, d(\pi, \sigma) = d(\tau\pi, \tau\sigma)$.

While this property may not be true of the Lovász Bregman divergences in general, the following theorem shows that this is true for a large class of them.

**Theorem 3.1.** *Given a submodular function $f$, such that $\forall \sigma, \tau \in \Sigma, h_{\tau\sigma}^f = \tau h_\sigma^f, d_{\hat{f}}(x||\sigma) = d_{\hat{f}}(\tau x||\tau\sigma)$.*

This property seems a little demanding for a submodular function. But a large class of submodular functions can be seen to have this property. In fact, it can be verified that any cardinality based submodular function has this property.

**Corollary 3.1.1.** *Given a submodular function $f$ such that $f(X) = g(|X|)$ for some function $g$, then $d_{\hat{f}}(x||\sigma) = d_{\hat{f}}(\tau x, \tau\sigma)$.*

This follows directly from Eqn. (11) and observing that the extreme points of the corresponding polyhedron are reorderings of each other. In other words, in these cases the submodular polyhedron forms a permutahedron. This property is true even for sums of such functions and therefore for many of the special cases which we have considered.

**Dependence on the values and not just the orderings:** We shall here analyze one key property of the LB divergence that is not present in other permutation based divergences. Consider the problem of combining rankings where, given a collection of scores $x^1, \cdots, x^n$, we want to come up with a joint ranking. An extreme case of this is where for some $x$ all the elements are the same. In this case $x$ expresses no preference in the joint ranking. Indeed it is easy to verify that for such an $x$, $d_{\hat{f}}(x||\sigma) = 0, \forall \sigma$. Now given a $x$ where all the elements are almost equal (but not exactly equal), even though this vector is totally ordered, it expresses a very low confidence in it's ordering. We would expect for such an $x$, $d_{\hat{f}}(x||\sigma)$ to be small for every $\sigma$. Indeed we have the following result:

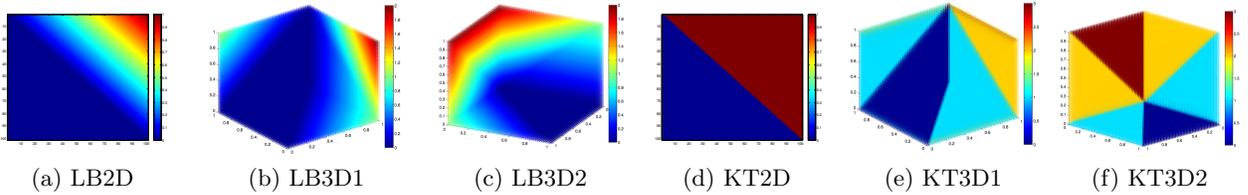

| (a) LB2D | (b) LB3D1 | (c) LB3D2 | (d) KT2D | (e) KT3D1 | (f) KT3D2 |

Figure 1: A visualization of $d_{\hat{f}}(x||\sigma)$ (left three) and $d_T(\sigma_x, \sigma)$ (right three). The figures shows a visualization in 2D, and two views in 3D for each, with $\sigma$ as $\{1, 2\}$ and $\{1, 2, 3\}$ and $x \in [0, 1]^2$ and $[0, 1]^3$ respectively.

**Theorem 3.2.** *Given a monotone submodular function $f$ and any permutation $\sigma$,*

$$d_{\hat{f}}(x||\sigma) \leq \epsilon n(\max_j f(j) - \min_j f(j|V \setminus j)) \leq \epsilon n \max_j f(j)$$

*where $\epsilon = \max_{i,j} |x_i - x_j|$ and $f(j|A) = f(A \cup j) - f(A)$.*

The above theorem implies that if the vector $x$ is such that all it's elements are almost equal, then $\epsilon$ is small and the LB divergence is also proportionately small. This bound can be improved in certain cases. For example for the cut function, with $f(X) = |X||V \setminus X|$, we have that $d_{\hat{f}}(x||\sigma) \leq \epsilon d_T(\sigma_x, \sigma) \leq \epsilon n(n-1)/2$, where $d_T$ is the Kendall $\tau$.

**Priority for higher rankings:** We show yet another nice property of the LB divergence with respect to a natural priority in rankings. This property has to do intrinsically with the submodularity of the generator function. We have the following theorem, that demonstrates this:

**Lemma 3.1.** *Given permutations $\sigma, \pi$, such that $\mathcal{P}(\sigma)$ and $\mathcal{P}(\pi)$ share a face (say $S_k^\sigma \neq S_k^\pi$) and $x \in \mathcal{P}(\pi)$), then $d_{\hat{f}}(x||\sigma) = (x_k - x_{k+1})(f(\sigma_x(k)|S_{k-1}^\sigma) - f(\sigma_x(k)|S_k^\sigma))$.*

This result directly follows from the definitions. Now consider the class of submodular function $f$ such that $\forall j, k \notin X, j \neq k, f(j|S) - f(j|S \cup k)$ is monotone decreasing as a function of $S$. An example of such a submodular function is again $f(X) = g(|X|)$, for a concave function $g$. Then it is clear that from the above Lemma that $d_{\hat{f}}(x||\sigma)$ will be larger for smaller $k$. In other words, if $\pi$ and $\sigma$ differ in the starting of the ranking, the divergence is more than if $\pi$ and $\sigma$ differ somewhere towards the end of the ranking. This kind of weighting is more prominent for the class of functions which depend on the cardinality, i.e., $f(X) = g(|X|)$. Recall that many of our special cases belong to this class. Then we have that $d_{\hat{f}}(x||\sigma) = \sum_{i=1}^n \{x(\sigma_x(i)) - x(\sigma(i))\} \delta_g(i)$. Now since $\delta_g(1) \geq \delta_g(2) \geq \cdots \geq \delta_g(n)$, it then follows that if $\sigma_x$ and $\sigma$ differ in the start of the ranking, they are penalized more.

**Extensions to partial orderings and top $m$-Lists:** So far we considered notions of distances between a score $x$ and a complete permutation $\sigma$. Often we may not be interested in a distance to a total ordering $\sigma$, but just a distance to a say a top-$m$ list [26] or a partial ordering between elements [38, 8]. The LB divergence also has a nice interpretation for both of these. In particular, in the context of top $m$ lists, we can use Eqn. (12). This exactly corresponds to the divergence between different or possibly overlapping sets of $m$ objects. Moreover, if we are simply interested in the top $m$ elements without the orderings, we have Eqn. (13). A special case of this is when we may just be interested in the top value. Another interesting instance is of partial orderings, where we do not care about the total ordering. For example, in web ranking we often just care about the relevant and irrelevant documents and that the relevant ones should be placed above the irrelevant ones. We can then define a distance $d_{\hat{f}}(x||\mathcal{P})$ where $\mathcal{P}$ refers to a partial ordering by using the cut based Lovász Bregman (Eqn. (10)) and defining the graph to have edges corresponding to the partial ordering. For example if we are interested in a partial order $1 > 2, 3 > 2$ in the elements $\{1, 2, 3, 4\}$, we can define $d_{1,2} = d_{3,2} = 1$ with the rest $d_{ij} = 0$ in Eqn. (10). Defined in this way, the LB divergence then measures the distortion between a vector $x$ and the partial ordering $1 > 2, 3 > 2$. In all these cases, we see that the extensions to partial rankings are natural in our framework, without needing to significantly change the expressions to admit these generalizations.

**Lovász-Mallows model:** In this section, we extend the notion of Mallows model to the LB divergence. We first define the Mallows model for the LB divergence:

$$p(x|\theta, \sigma) = \frac{\exp(-\theta d_{\hat{f}}(x||\sigma))}{Z(\theta, \sigma)}, \quad \theta \geq 0. \quad (16)$$

For this distribution to be a valid probability distribution, we assume that the domain $\mathcal{D}$ of $x$ to be a bounded set (say for example $[0, 1]^n$). We also assume that the domain is symmetric over permutations (i.e., for all $\sigma \in \Sigma$, if $x \in \mathcal{D}, x\sigma \in \mathcal{D}$. Unlike the standard Mallow's model, however, this is defined over scores (or valuations) as opposed to permutations.

Given the class of LB divergences defining a probability distribution over such a symmetric set (i.e., the divergences are invariant over relabelings) it follows that

$Z(\theta, \sigma) = Z(\theta)$. The reason for this is:

$$Z(\theta, \sigma) = \int_x \exp(-\theta d_{\hat{f}}(x, \sigma)) dx$$
$$= \int_x \exp(-\theta d_{\hat{f}}(x\sigma^{-1}, \sigma_0)) dx$$
$$= \int_{x'} \exp(-\theta d_{\hat{f}}(x', \sigma_0)) dx' = Z(\theta)$$

where $\sigma_0 = \{1, 2, \cdots, n\}$. We can also define an extended Mallows model for combining rankings, analogous to [27]. Unlike the Mallows model however this is a model over permutations given a collection of vectors $\mathcal{X} = \{x_1, \cdots, x_n\}$ and parameters $\Theta = \{\theta_1, \cdots, \theta_n\}$.

$$p(\sigma|\Theta, \mathcal{X}) = \frac{\exp(-\sum_{i=1}^n \theta_i d_{\hat{f}}(x_i||\sigma))}{Z(\Theta, \mathcal{X})} \quad (17)$$

This model can be used to combine rankings using the LB divergences, in a manner akin to Cranking [27]. This extended Lovász-Mallows model also admits an interesting Bayesian interpretation, thereby providing a generative view to this model:

$$p(\sigma|\Theta, \mathcal{X}) \propto p(\sigma) \prod_{i=1}^n p(x_i|\sigma, \theta_i). \quad (18)$$

Again this directly follows from the fact that in this case, in the Lovász-Mallows model, the normalizing constants (which are independent of $\sigma$) cancel out. We shall actually see some very interesting connections between this conditional model and web ranking.

## 4 Applications

**Rank Aggregation:** As argued above, the LB divergence is a natural model for the problem of combining scores, where both the ordering and the valuations are provided. If we ignore the values, but just consider the rankings, this then becomes rank aggregation. A natural choice in such problems is the Kendall $\tau$ distance [27, 26, 31]. On the other hand, if we consider only the values without explicitly modeling the orderings, then this becomes an incarnation of boosting [16]. The Lovász-Bregman divergence tries to combine both aspects of this problem – by combining orderings using a permutation based divergence, while simultaneously using the additional information of the confidence in the orderings provided by the valuations. We can then pose this problem as:

$$\sigma \in \underset{\sigma' \in \Sigma}{\operatorname{argmin}} \sum_{i=1}^n d_{\hat{f}}(x^i||\sigma') \quad (19)$$

The above notion of the representative ordering (also known as the mean ordering) is very common in many applications [3] and has also been used in the context of combining rankings [31, 27, 26]. Unfortunately this problem in the context of the permutation based metrics were shown to be NP hard [4]. Surprisingly for the LB divergence this problem is easy (and has a closed form). In particular, the representative permutation is exactly the ordering corresponding to the arithmetic mean of the elements in $\mathcal{X}$.

**Lemma 4.1.** *[19] Given a submodular function $f$, the Lovász Bregman representative (Eqn. (19)) is $\sigma = \sigma_\mu$, where $\mu = \frac{1}{n} \sum_{i=1}^n x^i$*

This result builds on the known result for Bregman divergences [3]. This seems somewhat surprising at first. Notice, however, that the arithmetic mean uses additional information about the scores and its confidence, as opposed to just the orderings. In this context, the result then seems reasonable since we would expect that the representatives be closely related to the ordering of the arithmetic mean of the objects. We shall also see that this notion has in fact been ubiquitously but unintentionally used in the web ranking and information retrieval communities.

We illustrate the utility of the Lovász Bregman rank aggregation through the following argument. Assume that a particular vector $x$ is uninformative about the true ordering (i.e, the values of $x$ are almost equal). Then with the LB divergence and any permutation $\pi$, $d(x||\pi) \approx 0$, and hence this vector will not contribute to the mean ordering. Instead if we use a permutation based metric, it will ignore the values but consider only the permutation. As a result, the mean ordering tends to consider such vectors $x$ which are uninformative about the true ordering. As an example, consider a set of scores: $\mathcal{X} = \{1.9, 2\}, \{1.8, 2\}, \{1.95, 2\}, \{2, 1\}, \{2.5, 1.2\}$. The representative of this collection as seen by a permutation based metric would be the permutation $\{1, 2\}$ though the former three vectors have very low confidence. The arithmetic mean of these vectors is however $\{2.03, 1.64\}$ and the Lovász Bregman representative would be $\{2, 1\}$.

The arithmetic mean also provides a notion of confidence of the population. In particular, if the total variation [2] of the arithmetic mean is small, it implies that the population is not confident about its ordering, while if the variation is high, it provides a certificate of a homogeneous population. Figure 1 provides a visualization the Lovász-Bregman divergence using the cut function and the Kendall $\tau$ metric, visualized in 2 and 3 dimensions respectively. We see the similarity between the two divergences and at the same time, the dependence on the "scores" in the Lovász-Bregman case.

**Learning to Rank:** We investigate a specific instance of the rank aggregation problem with reference to the problem of "learning to rank." A large class of algorithms have been proposed for this problem – see [28] for a survey on this. A specific class of algorithms for this problem have focused on maximum margin learning using ranking based loss functions

(see [38, 8] and references therein). While we have seen that the ranking based losses themselves are instances of the LB divergence, the feature functions are also closely related.

In particular, given a query $q$, we denote a feature vector corresponding to document $i \in \{1, 2, \cdots, n\}$ as $x_i \in \mathbb{R}^d$, where each element of $x_i$ denotes a quality of document $i$ based on a particular indicator or feature. Denote $\mathcal{X} = \{x_1, \cdots, x_n\}$. We assume we have $d$ feature functions (one might be for example a match with the title, another might be pagerank, etc). Denote $x_i^j$ as the score of the $j^{\text{th}}$ feature corresponding to document $i$ and $x^j \in \mathbb{R}^n$ as the score vector corresponding to feature $j$ over all the documents. In other words, $x^j = (x_1^j, x_2^j, \cdots, x_n^j)$. One possible choice of feature function is:

$$\phi(\mathcal{X}, \sigma) = \sum_{j=1}^{d} w_j d_{\hat{f}}(x^j || \sigma) \qquad (20)$$

for a weight vector $w \in \mathbb{R}^d$. Given a particular weight vector $w$, the inference problem then is to find the permutation $\sigma$ which minimizes $\phi(\mathcal{X}, \sigma)$. Thanks to Lemma 4.1, the permutation $\sigma$ is exactly the ordering of the vector $\sum_{j=1}^{n} w_j x^j$. It is not hard to see that this exactly corresponds to ordering the scores $w^\top x_i$ for $i \in \{1, 2, \cdots, n\}$. Interestingly many of the feature functions used in [38, 8] are forms closely related Eqn. (20). In fact the motivation to define these feature functions is exactly that the inference problem for a given set of weights $w$ be solved by simply ordering the scores $w^\top x_i$ for every $i \in \{1, 2, \cdots, n\}$ [8]. We see that through Eqn. (20), we have a large class of possible feature functions for this problem.

We also point out a connection between the learning to rank problem and the Lovász-Mallows model. In particular, recent work [12] defined a conditional probability model over permutations as:

$$p(\sigma | w, \mathcal{X}) = \frac{\exp(w^\top \phi(\mathcal{X}, \sigma))}{Z}. \qquad (21)$$

This conditional model is then exactly the extended Lovász-Mallows model of Eqn. (17) when $\phi$ is defined as in Eqn. (20). The conditional models used in [12] are in fact closely related to this and correspondingly Eqn. (21) offers a large class of conditional ranking models for the learning to rank problem.

**Clustering:** A natural generalization of rank aggregation is the problem of clustering. In this context, we assume a heterogeneous model, where the data is represented as mixtures of ranking models, with each mixture representing a homogeneous population. It is natural to define a clustering objective in such scenarios. Assume a set of representatives $\Sigma = \{\sigma^1, \cdots, \sigma^k\}$ and a set of clusters $\mathcal{C} = \{\mathcal{C}^1, \mathcal{C}^2, \cdots, \mathcal{C}^k\}$. The clustering objective is then: $\min_{\mathcal{C}, \Sigma} \sum_{j=1}^{k} \sum_{i: x_i \in \mathcal{C}_j} d_{\hat{f}}(x_i || \sigma_i)$. As shown in [19], a simple k-means style algorithm finds a local minima of the above objective. Moreover due to simplicity of obtaining the means in this case, this algorithm is extremely scalable and practical.

## 5 Discussion

To our knowledge, this work is the first introduces the notion of "score based divergences" in preference and ranking based learning. Many of the results in this paper are due to some interesting properties of the Lovász extension and Bregman divergences. This also provides interesting connections between web ranking and the permutation based metrics. This idea is mildly related to the work of [36] where they use the Choquet integral (of which the Lovász extension is a special case) for preference learning. Unlike our paper, however, they do not focus on the divergences formed by the integral. Finally, it will be interesting to use these ideas in real world applications involving rank aggregation, clustering, and learning to rank.

**Acknowledgments:** We thank Matthai Phillipose, Stefanie Jegelka and the melodi-lab submodular group at UWEE for discussions and the anonymous reviewers for very useful reviews. This material is based upon work supported by the National Science Foundation under Grant No. (IIS-1162606), and is also supported by a Google, a Microsoft, and an Intel research award.